\documentclass[letterpaper]{article}

\makeatletter
\def\bibliographystyle#1{}%
\def\bibliography#1{}%
\long\def\pdfinfo#1{\relax}%
\makeatother

\newif\ifaaaistyle
\IfFileExists{aaai2026.sty}{\aaaistyletrue}{\aaaistylefalse}

\ifaaaistyle
  \usepackage{aaai2026}
  \nocopyright
\else
  \usepackage[margin=1in]{geometry}
\fi

\usepackage{times}  
\usepackage{helvet}  
\usepackage{courier}  

\usepackage[hyphens]{url}  
\usepackage{graphicx} 
\usepackage{natbib}  
\usepackage{caption} 
\ifdefined\pdfpagewidth
\else
\fi
\usepackage{algorithm}
\usepackage{algorithmic}
\usepackage{multirow}
\usepackage{makecell}

\usepackage{newfloat}
\usepackage{listings}
\DeclareCaptionStyle{ruled}{labelfont=normalfont,labelsep=colon,strut=off} 
\floatstyle{ruled}
\newfloat{listing}{tb}{lst}{}
\floatname{listing}{Listing}
\usepackage{booktabs}
\usepackage{array}
\usepackage[table]{xcolor}
\usepackage{amsmath}
\usepackage{amssymb}
\usepackage{amsthm}

\newtheorem{definition}{Definition}[section]
\usepackage[most]{tcolorbox}

\makeatletter
\renewcommand\paragraph{\@startsection{paragraph}{4}{\z@}%
  {0.55ex \@plus 0.25ex \@minus 0.1ex}%
  {0.2ex}%
  {\normalsize\bfseries}}
\renewcommand\normalsize{%
  \@setfontsize\normalsize\@xpt{12}%
  \abovedisplayskip 10\p@ \@plus2\p@ \@minus5\p@
  \abovedisplayshortskip \z@ \@plus3\p@
  \belowdisplayshortskip 6\p@ \@plus3\p@ \@minus3\p@
  \belowdisplayskip \abovedisplayskip
  \let\@listi\@listI}
\normalsize
\makeatother

\makeatletter
\def\bibliographystyle#1{}%
\def\bibliography#1{}%
\AtBeginDocument{%
  \def\bibliographystyle#1{}%
  \def\bibliography#1{}%
}
\makeatother

\title{SodaMem: Evidence-Grounded Temporal Graph Memory for LLM Agents}

\author{
Fengrong Wan\thanks{Corresponding author. Email: xlows1206@gmail.com},
Chengcan Wu\textsuperscript{\rm 1},
Ningtao Lyu\\
}
\ifaaaistyle
\affiliations{
\textsuperscript{\rm 1}Peking University
}
\fi

\begin{document}
\maketitle

\begin{abstract}
Large language model (LLM) agents that assist users over weeks of conversation must remember \emph{what is currently true}, not merely \emph{what was once said}. Flat RAG diaries and Markdown logs optimize needle retrieval but under-serve currency, provenance, and ordered temporal reasoning~\cite{locomo2024,longmemeval2024,memgpt2023,mem02025}. We present \textbf{SodaMem}, an evidence-grounded \emph{temporal graph memory} that (i)~extracts typed FactEvents with mandatory provenance spans, (ii)~persists mention time, occurrence time, and validity with $\textsc{SUPERSEDES}$/$\textsc{CONTRADICTS}$/$\textsc{UPDATES}$ edges under hybrid lexical--dense indexing, and (iii)~answers via a planner--reader loop that gathers citable evidence before composing a final response. On \textbf{LongMemEval-S}, our store-of-record configuration reaches \textbf{92.8\%} accuracy ($464/500$; best of $N{=}3$) at mean \$0.00161/question ($\approx$18.3k tokens; median \$0.00111 / $\approx$14.6k) with \texttt{deepseek-v4-flash}. We compile public systems with estimable API cost into a cost table and cost--accuracy map; under these estimates SodaMem sits near the accuracy frontier at Flash-tier spend and strictly dominates several higher-cost, lower-accuracy points. Accuracy uses the same Flash model as reader and judge (self-grading); costs exclude ingest/judge and cross-system comparisons are compiled estimates rather than a single-harness bake-off. Our code is available at \url{https://github.com/SodaMem/SodaMem}.
\end{abstract}

\section{Introduction}

LLM agents that accompany users across days and weeks are routinely given a ``memory'': append-only chat logs, Markdown diaries, vector stores, summarization pipelines, or increasingly elaborate graph and hierarchy designs~\cite{locomo2024,longmemeval2024,memgpt2023,memorybank2023,amem2025,zep2025,mem02025,memorysurvey2026}. Despite rapid growth of this literature~\cite{zhang2024memorysurvey,xi2023survey,wang2024agentsurvey}, the \emph{research focus} of long-horizon personal memory has consolidated around a small set of measurable pressures rather than a single architecture. Benchmarks such as LoCoMo and LongMemEval probe multi-session fact recall, knowledge updates, temporal reasoning, preference tracking, and abstention~\cite{locomo2024,longmemeval2024}; complementary suites stress implicit state invalidation~\cite{stale2026}, memory-operation correctness~\cite{memops2026}, prospective triggering~\cite{pmbench2026}, MemBench-style axes~\cite{membench2025}, and agent--environment experience beyond chat~\cite{longmemevalv2,memoryarena2026,memoryagentbench2025}. In short, the field is no longer asking only ``can the model find a needle in the transcript?''; but ``can the agent maintain a coherent, updatable model of the user (or environment) and \emph{use} it under the right conditions?''

\paragraph{Method focus and open problems.}
Systems work spans a familiar pipeline---land, structure, index, link, maintain, retrieve/answer~\cite{agentnative2026,memorysurvey2026,lewis2020rag,amem2025,zep2025,simplemem2026,memoryr1,napmem2026,mem02025}---but everyday assistants still hit four failure modes. \textbf{(P1) Currency / conflict:} preferences reverse; append-only logs leave ``which value is current?'' to an LLM over unordered chunks, where deterministic freshness often beats free-form judgment~\cite{detconflict2026,stale2026}. \textbf{(P2) Temporal structure:} ordering / ``most recently'' / relative-date questions break when relative phrases lack a comparable timeline. \textbf{(P3) Provenance:} citations to source turns are needed for trust; lossy summaries and opaque vector hits weaken audit. \textbf{(P4) Association:} multi-hop synthesis needs entity/claim links beyond cosine neighbors, while avoiding \emph{context collapse} from episode-wrong but similar memories~\cite{ramem2026}.

\begin{figure*}[t!]
\centering
\includegraphics[width=0.85\textwidth]{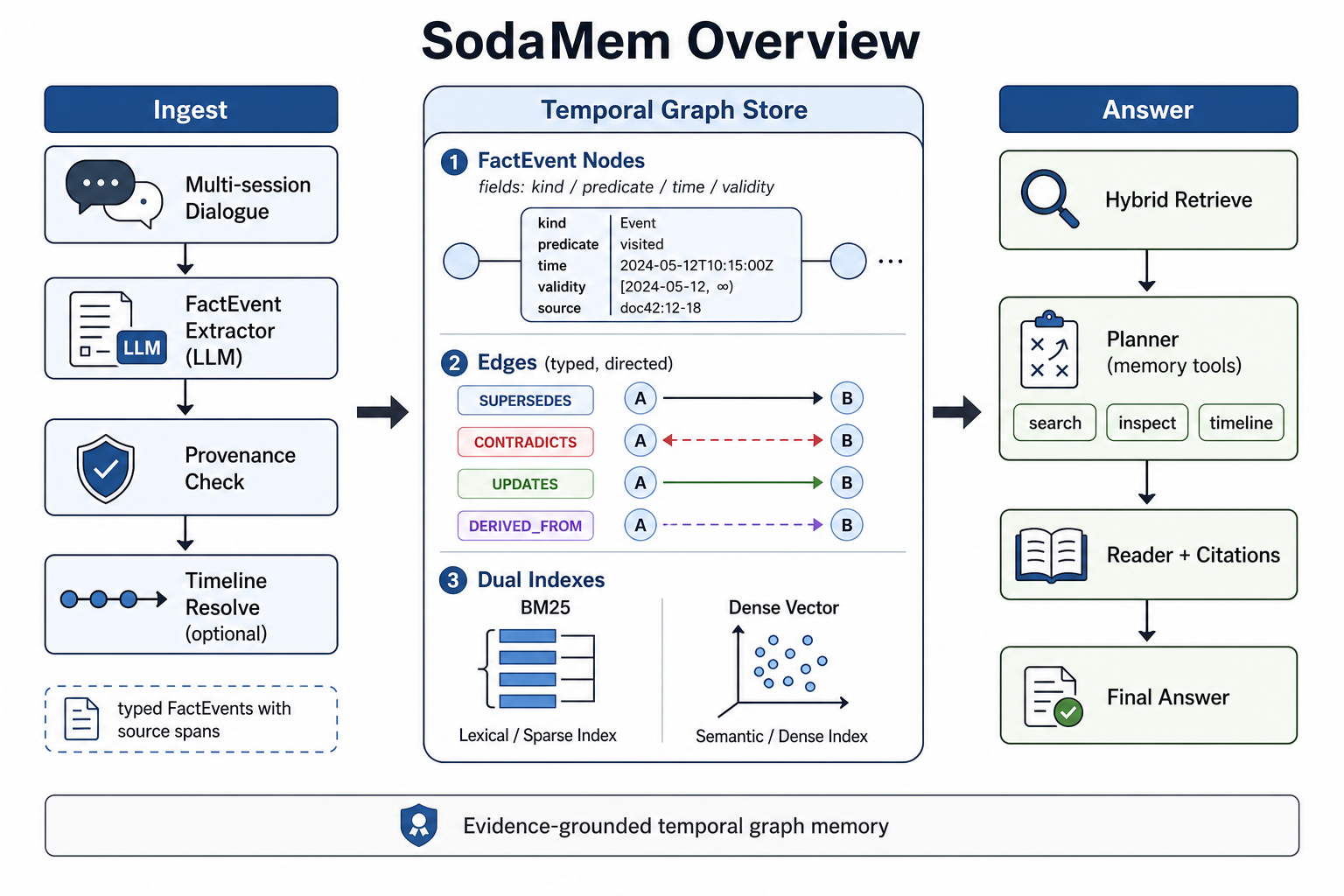}
\caption{Overview of SodaMem. Multi-session dialogue is ingested into typed FactEvents with provenance checks and optional timeline resolution. A temporal graph store holds FactEvent nodes (kind, predicate, time, validity), typed edges ($\textsc{SUPERSEDES}$, $\textsc{CONTRADICTS}$, $\textsc{UPDATES}$, $\textsc{DERIVED\_FROM}$ / semantic \& relation edges), and dual BM25 / dense indexes. At question time, multi-tunnel retrieval with connection-density fusion feeds a planner over memory tools; a separate reader produces a cited final answer.}
\label{fig:sodamem-overview}
\end{figure*}

\paragraph{Our stance.}
For retrospective personal QA we treat memory as an \textbf{evidence-grounded temporal knowledge graph}: typed FactEvents with source spans, temporal axes (mention, occurrence, validity), and $\textsc{SUPERSEDES}$/$\textsc{CONTRADICTS}$/$\textsc{UPDATES}$ edges; a planner--reader loop gathers evidence before prose. This complements RL controllers and prospective-memory suites: we prioritize a maintainable, citable state for LongMemEval-style questions, with timeline resolution for temporal misses.

We instantiate the stance with \textbf{SodaMem} (Figure~\ref{fig:sodamem-overview}):
\begin{enumerate}
    \item \textbf{Ingest:} LLM extraction of FactEvents with provenance hard constraints and modality/calendar post-processing; optional entity-subject prompts to reduce star-graph collapse.
    \item \textbf{Store \& maintain:} SQLite facts plus hybrid BM25--dense indexes; supersession and contradiction edges; dream/maintenance; optional session-anchored timeline resolution for relative dates.
    \item \textbf{Answer:} hybrid recall, a multi-step planner over memory tools, and a separate reader that emits cited answers.
\end{enumerate}

\paragraph{How SodaMem addresses (P1)--(P4).}
Supersession and validity closing target (P1); timeline resolution and temporal fields on FactEvents target (P2); mandatory source spans and reader citations target (P3); typed predicates, entity roles, and graph edges target (P4), while the planner can expand sessions and inspect cards to reduce episode confusion relative to single-shot RAG.

\paragraph{Contributions.}
\begin{itemize}
    \item \textbf{Problem framing.} We synthesize agent-memory research foci and method axes, and isolate currency, temporal structure, provenance, and association as the failure modes that Markdown/flat RAG leave unresolved for long-horizon personal assistants.
    \item \textbf{System.} We present SodaMem's ingest--store--planner--reader pipeline: FactEvent schema, hybrid retrieval, supersession semantics, and a proposed timeline-resolution layer aimed at temporal-reasoning errors.
    \item \textbf{Cost--accuracy evaluation.} On \textbf{LongMemEval-S} we report a store-of-record $92.8\%$ run at mean \$0.00161/question ($\approx$18.3k tokens; median \$0.00111 / $\approx$14.6k), compile public baselines with estimable API cost into a cost table and cost--accuracy map, and analyze the dominated (higher-cost, lower-accuracy) region relative to SodaMem.
\end{itemize}

\section{Related Work}

\paragraph{Benchmarks.}
LoCoMo~\cite{locomo2024} and LongMemEval~\cite{longmemeval2024} are the main yardsticks for retrospective personal-memory QA (multi-session recall, updates, temporal reasoning, preference, abstention). Broader suites probe implicit invalidation, memory-operation correctness, prospective triggering, and agent--environment experience~\cite{stale2026,memops2026,pmbench2026,membench2025,memoryagentbench2025,memoryarena2026,longmemevalv2}. We evaluate on LongMemEval-S and treat the others as orthogonal pressures.

\paragraph{External memory and structure.}
RAG and long-context readers~\cite{lewis2020rag,karpukhin2020dpr} serve static corpora; agent settings continually write user state. MemGPT-style paging~\cite{memgpt2023}, MemoryBank / hierarchical summarization~\cite{memorybank2023,lee2024readagent}, and Mem0-style extractive APIs~\cite{mem02025} establish the need for an external store. Hierarchical and note/graph designs~\cite{napmem2026,lightmem2026,amem2025,zep2025,graphrag2024,hipporag2024} move beyond flat chunks via compression, Zettelkasten links, or bi-temporal graphs with edge invalidation. SodaMem is closest to extraction-plus-temporal-graph lines, but requires provenance spans, mention/occurrence/validity axes, write-time supersession, and a cited planner--reader loop.

\paragraph{Indexing, conflict, and controllers.}
SimpleMem stresses density gating and multi-view indexes~\cite{simplemem2026}; RaMem highlights context collapse~\cite{ramem2026}; deterministic conflict work favors explicit version marks over free-form freshness judgment~\cite{detconflict2026}. Parallel lines learn memory-tool policies with RL~\cite{memoryr1,agemem2026} or optimize multi-turn search~\cite{searchr1}. Markdown diaries remain a strong simplicity baseline---cheap but weak on currency, order, provenance, and association. SodaMem is an engineering-first temporal graph substrate for retrospective personal QA; learned controllers can later sit on the same FactEvent contract.

\section{Motivation and Preliminaries}

\subsection{Motivation}

\paragraph{Currency and multi-signal recall.}
Consider ``I love spicy food,'' later ``I am cutting down on spice,'' then ``What should I cook tonight?'' A Markdown log keeps all three; a flat retriever may surface the first. A temporal graph should supersede (or validity-close) the old preference, answer from the latest state, and still cite the justifying turns---the pattern behind LongMemEval knowledge-update items~\cite{longmemeval2024,detconflict2026}. Even with correct facts stored, single-channel retrieval is brittle: embeddings can be episode-wrong~\cite{ramem2026}, BM25 may miss paraphrase, and entity expansion can explode. We therefore use \textbf{wide multi-signal recall} (graph, BM25, dense) and rank by connection density across auditable links, not cosine alone.

\paragraph{Soft time and design principles.}
Users often misremember windows (``two months ago'' for a three-month-old fact); hard temporal filters then drop the right evidence. We store comparable temporal fields, parse query time into a soft window plus sort direction, and treat window match as a \textbf{bonus} in ranking confidence. Principles:
\begin{enumerate}
    \item \textbf{Evidence first:} no durable claim without a source span.
    \item \textbf{Explicit time:} mention, occurrence, and validity; query $\rightarrow$ window + sort direction.
    \item \textbf{Writable currency:} supersession/contradiction are first-class; invalid facts are excluded.
    \item \textbf{Multi-signal wide recall} with per-head expansion, then fusion.
    \item \textbf{Connection-density ranking} with soft time bonuses and near-duplicate merge.
    \item \textbf{Toolful answering:} planner--reader gather-then-read~\cite{memoryintheloop2026}.
\end{enumerate}

\subsection{Preliminaries}

\begin{definition}[FactEvent]
A FactEvent is $f=(\kappa,\pi,m,\tau,\rho,S,\sigma)$: kind $\kappa$, predicate $\pi$, modality $m$, temporal fields $\tau$, entity roles $\rho$, source spans $S$ (MessagePieces), and status $\sigma$ (active / superseded / invalid).
\end{definition}

Retrieval units are FactEvents, MessagePieces, or raw turns (stable IDs for fusion). Temporal axes: \textbf{mention} (session time $t_s$), \textbf{occurrence} ($occurred\_start/end$), and \textbf{validity} ($valid\_from/until$), closed under supersession.

\begin{definition}[Evidence-grounded answer]
Answer $a$ is evidence-grounded if each material claim is supported by retrieved $E\subseteq\mathcal{M}$ with non-empty provenance $S(f)$ for $f\in E$, and citations name those records.
\end{definition}

\begin{definition}[Supersession]
$f_{\mathrm{new}}$ supersedes $f_{\mathrm{old}}$ on a competing subject--predicate slot (or matched update pattern); then $\sigma(f_{\mathrm{old}})$ becomes superseded and $valid\_until(f_{\mathrm{old}})$ closes at the effective time of $f_{\mathrm{new}}$.
\end{definition}

\begin{definition}[Query temporal intent]
A parser maps $q$ to $(\mathcal{W},\delta)$: window $\mathcal{W}$ and sort $\delta\in\{\mathrm{near{\to}far},\mathrm{far{\to}near}\}$. Absent cues, $\mathcal{W}=\emptyset$ and $\delta=\mathrm{near{\to}far}$.
\end{definition}

\section{Proposed Method: SodaMem}

SodaMem is a memory infrastructure for LLM agents comprising ingest, durable storage with hybrid multi-signal retrieval, optional maintenance (dream / timeline resolution), and a planner--reader answering loop (Figure~\ref{fig:sodamem-overview}). Relative to Markdown diaries and flat RAG, the design goal is a \emph{maintainable} user knowledge state that remains citable---and a retrieval stack that ranks evidence by \emph{connection density} across graph, lexical, and dense channels under soft temporal scoring.

\subsection{Problem Statement}

Given a user $u$, multi-session dialogue history $\mathcal{H}_u=\{H_s\}_{s=1}^{S}$ with session times $\{t_s\}$, and question $q$, produce an evidence-grounded answer $a$ maximizing judge agreement with gold $a^\star$. We factor the system as
\begin{align}
\mathcal{M}_u &= \mathrm{Ingest}(\mathcal{H}_u), \\
E &= \mathrm{Retrieve}(q,\mathcal{M}_u), \\
a &= \mathrm{Read}(q,E).
\end{align}
SodaMem specifies $\mathrm{Ingest}$, the schema of $\mathcal{M}_u$, $\mathrm{Retrieve}$ (multi-signal recall + density fusion + optional planner tools), and $\mathrm{Read}$.

\subsection{Ingest: From Turns to FactEvents}

\paragraph{Segmentation and extraction.}
For each session $H_s$ with time $t_s$, turns are segmented and passed to an extractor LLM that emits FactEvent candidates under a fixed schema: kind, predicates, modality, temporal expressions, entity roles, \texttt{source\_span\_ids}, and \texttt{support\_text}. Candidates must name spans that literally occur in the source turn (MessagePieces).

\paragraph{Provenance hard constraint.}
Candidates whose spans do not land in the source turn are rejected. Raw turns keep stable \texttt{rawTurn\_id}s so later BM25/embedding hits can carry full turn text for similarity, deduplication, and density accounting.

\paragraph{Deterministic post-processing.}
Post-steps normalize modality, resolve absolute dates when stated, and attach $t_s$ as the mention-time anchor. Optional \textbf{coarse} and \textbf{entity-subject} prompts control granularity and reduce star-graph collapse onto $entity\_user$.

\begin{algorithm}[t]
\caption{IngestSession($H_s$, $t_s$)}
\begin{algorithmic}[1]
\STATE $\mathcal{C} \leftarrow \mathrm{ExtractLLM}(H_s)$
\STATE $\mathcal{F} \leftarrow \emptyset$
\FOR{each candidate $c\in\mathcal{C}$}
  \IF{$\mathrm{SpansValid}(c, H_s)$}
    \STATE $c \leftarrow \mathrm{NormalizeModalityAndDates}(c, t_s)$
    \STATE $\mathcal{F} \leftarrow \mathcal{F}\cup\{c\}$
  \ENDIF
\ENDFOR
\STATE $\mathcal{F} \leftarrow \mathrm{TimelineResolve}(\mathcal{F}, t_s)$ \COMMENT{optional}
\STATE $\mathrm{WriteFactsAndEdges}(\mathcal{F})$
\STATE \textbf{return} $\mathcal{F}$
\end{algorithmic}
\end{algorithm}

\subsection{Timeline Resolution Layer}

Relative phrases at \emph{ingest} are under-specified if left only as text. We optionally apply
\begin{equation}
\hat{\tau}(f) = \mathcal{T}\big(\tau_{\mathrm{raw}}(f),\, t_s(f),\, \mathrm{context}(f)\big),
\end{equation}
producing comparable timestamps. Unresolvable cases are marked \texttt{unresolved}. At \emph{query} time, a separate parser yields temporal intent $(\mathcal{W},\delta)$ used in soft temporal scoring below---aligned with bi-temporal / episodic concerns~\cite{zep2025,ramem2026}, but coupled to density fusion rather than hard episode filters alone.

\subsection{Store: Hybrid Index and Graph Relations}

\paragraph{Persistence.}
Facts persist in SQLite with dense vectors (MiniLM / GTE-class) and BM25 over fact text, spans, and raw turns. Cards expose predicate text, temporal fields, entity roles, status, and provenance.

\paragraph{Edges.}
We maintain mention / $\textsc{DERIVED\_FROM}$ links to spans; $\textsc{SUPERSEDES}$ / $\textsc{CONTRADICTS}$ / $\textsc{UPDATES}$ among facts; and graph expansion edges of two flavors used at retrieve time: \textbf{semantic edges} (content-driven neighbor links) and \textbf{relation-type edges} (typed predicates between entities). Product defaults write supersession; observe-only frozen stores are an experimental axis. Dreaming rebuilds dirty entity profiles.

\subsection{Retrieve: Multi-Tunnel Recall and Connection-Density Fusion}

Retrieval is the core of $\mathrm{Retrieve}(q,\mathcal{M}_u)$ and follows the initial SodaMem design: \textbf{wide multi-path recall}, per-tunnel head expansion, validity gates, then fusion by connection density with soft time bonuses.

\paragraph{Query analysis.}
Parse $q$ into entity mentions, lexical keys, an embedding query, and temporal intent $(\mathcal{W},\delta)$. Vague cues (``recently'', ``a few months ago'') are mapped to \emph{wide} windows to prefer recall over precision; missing cues disable the time bonus rather than inventing a window.

\paragraph{Three tunnels (strong vs.\ weak).}
\begin{itemize}
    \item \textbf{Graph / entity tunnel (strong):} hit entities or facts as \emph{search heads}; expand along selected semantic or relation-type edges (1-hop or limited multi-hop). Each head expands independently, then applies validity, relevance, and soft time scoring; keep \texttt{search\_head\_rerank\_top\_K}.
    \item \textbf{BM25 tunnel (strong):} lexical hits on facts, MessagePieces, or raw turns. Span hits attach neighboring spans and the parent \texttt{rawTurn} (full text as a field for similarity/dedup); raw-turn hits expand to temporally adjacent turns ($\pm 2$) for local context.
    \item \textbf{Embedding tunnel (weak):} dense neighbors with the same expansion patterns as BM25, but lower base weights because similarity may retrieve related-but-irrelevant episodes.
\end{itemize}
Each tunnel uses at most $H$ search heads (default $H{=}10$). Direct hits from strong tunnels receive higher base mass than weak-tunnel or derived (expanded) hits.

\paragraph{Validity gate (hard).}
Exclude content whose status is invalid/superseded-as-inactive when inappropriate, or whose validity interval is incompatible with $\mathcal{W}$ when a window is stated \emph{and} the fact's validity is known. This is the only hard temporal/status exclusion; occurrence-time mismatch alone does not drop a high-density candidate.

\paragraph{Connection density and ranking confidence.}
Let each (tunnel, head, hit) award a mass $w$ to an evidence ID $i$ (fact / span / rawTurn). Defaults (tunable): strong direct $0.4$, weak direct $0.2$, strong derived $0.1$, weak derived $0.05$. Masses \textbf{accumulate} when multiple heads hit the same ID (by ID equality, or by embedding similarity $\ge\theta$, e.g.\ $0.8$, for near-duplicate merge). Writing $H(i)$ for the hits on $i$,
\begin{align}
\mathrm{density}(i) &= \sum_{h\in H(i)} w_h, \\
\mathrm{conf}(i) &= \mathrm{density}(i)
  + \beta\cdot\mathbf{1}[i\cap\mathcal{W}\neq\emptyset],
\end{align}
with time bonus $\beta$ (default $0.3$) awarded at most once per merged item if any constituent falls in $\mathcal{W}$. Sort by $\mathrm{conf}$ (ties broken by $\delta$). Time thus acts as a ranking feature rather than a hard filter, so user-misdated queries remain recoverable.

\paragraph{Fusion.}
Merge per-tunnel lists by ID/similarity, recompute $\mathrm{conf}$, and emit a unique ranked pool (Recall@$k$ cutoffs are experimental knobs).

\begin{algorithm}[t]
\caption{MultiTunnelRetrieve($q$, $\mathcal{M}_u$)}
\begin{algorithmic}[1]
\STATE $(\mathcal{W},\delta)\leftarrow\mathrm{ParseTemporal}(q)$
\STATE $\textit{pools}\leftarrow\emptyset$
\FOR{tunnel $t\in\{\mathrm{graph},\mathrm{BM25},\mathrm{embed}\}$}
  \STATE $\textit{heads}\leftarrow\mathrm{TopHeads}(t,q,\mathcal{M}_u;H)$
  \FOR{head $h\in\textit{heads}$}
    \STATE $L\leftarrow\mathrm{Expand}(h,t)$;\; $L\leftarrow\mathrm{ValidityFilter}(L,\mathcal{W})$
    \STATE $L\leftarrow\mathrm{RerankLocal}(L,q,\mathcal{W},\delta;\texttt{top\_K})$
    \STATE $\textit{pools}\leftarrow\textit{pools}\cup\mathrm{AwardMass}(L,t)$
  \ENDFOR
\ENDFOR
\STATE $E\leftarrow\mathrm{MergeByIdOrSim}(\textit{pools};\theta)$;\; score $\mathrm{conf}$ on $E$
\STATE \textbf{return} top evidence by $\mathrm{conf}$ under $\delta$
\end{algorithmic}
\end{algorithm}

\subsection{Answer: Planner--Reader Loop}

\paragraph{Planner.}
An LLM may further call tools (\texttt{search}, \texttt{inspect}, \texttt{session\_expand}, \texttt{timeline}, \texttt{count}, \texttt{compute}) under a step budget to grow the fused pool---implementing memory-in-the-loop~\cite{memoryintheloop2026} when density ranking alone is insufficient (e.g., explicit enumeration).

\paragraph{Reader.}
A separate prompt composes the user-facing answer from selected evidence IDs with mandatory citations. Separation keeps citation discipline out of the tool policy.

\begin{algorithm}[t]
\caption{Answer($q$, $\mathcal{M}_u$)}
\begin{algorithmic}[1]
\STATE $E \leftarrow \mathrm{MultiTunnelRetrieve}(q, \mathcal{M}_u)$
\STATE $\textit{open} \leftarrow \{q\}$
\FOR{$t = 1$ to $T_{\max}$}
  \STATE $\textit{act} \leftarrow \mathrm{Planner}(q, E, \textit{open})$
  \IF{$\textit{act}=\textsc{stop}$} \STATE \textbf{break} \ENDIF
  \STATE $E \leftarrow E \cup \mathrm{ExecTool}(\textit{act}, \mathcal{M}_u)$
\ENDFOR
\STATE $a \leftarrow \mathrm{Reader}(q, E)$
\STATE \textbf{return} $a$ with citations into $E$
\end{algorithmic}
\end{algorithm}

\subsection{Implementation Notes}

Frozen LongMemEval stores open read-only with fingerprint echo. Density weights $(0.4,0.2,0.1,0.05)$, $\beta$, $\theta$, $H$, and \texttt{search\_head\_rerank\_top\_K} are exposed for Recall@$k$ sweeps.

\section{Experiments}

\begin{figure*}[t]
\centering
\includegraphics[width=0.92\textwidth]{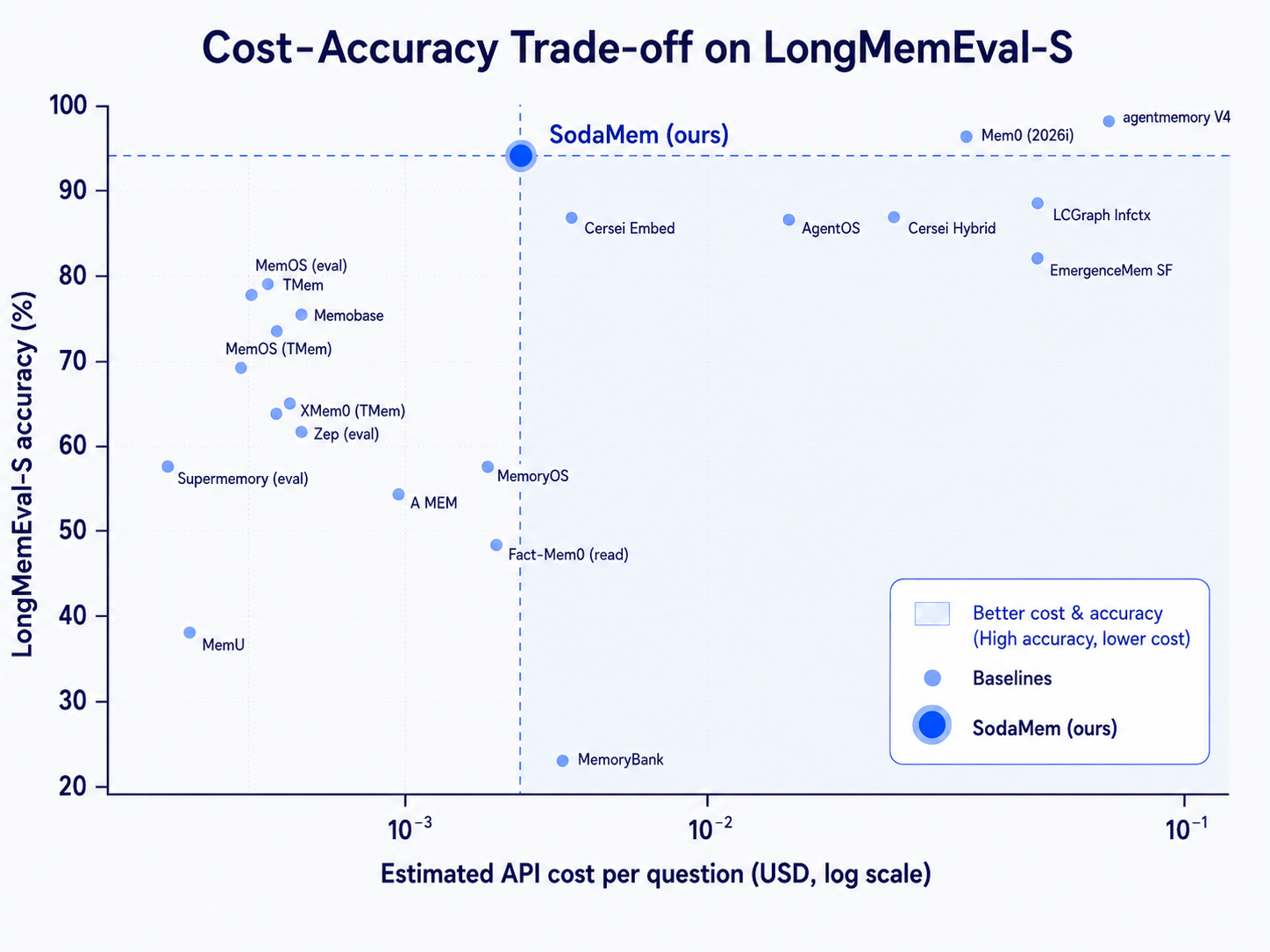}
\caption{LongMemEval-S accuracy vs.\ estimated API cost per question (log $x$-axis). SodaMem is the star (mean cost \$0.00161; median \$0.00111 / $\approx$14.6k tokens). Blue dashed lines mark our \emph{mean} cost and $92.8\%$ accuracy; the shaded quadrant is strictly dominated by SodaMem (higher cost and lower accuracy than the mean operating point).}
\label{fig:cost-acc}
\end{figure*}

We evaluate SodaMem on \textbf{LongMemEval-S} (500 questions; $\approx$115k-token histories)~\cite{longmemeval2024} via the \emph{accuracy--cost} trade-off against publicly reported systems with estimable per-question API cost. We compile disclosed scores, models, and token/\$ figures from primary sources (rather than re-running every baseline under one harness), convert them with 2026 list prices, and situate our store-of-record run in that landscape. Table~\ref{tab:lme-cost} sorts methods by accuracy; Figure~\ref{fig:cost-acc} plots the same points.

\begin{table*}[t]
\centering
\scriptsize
\setlength{\tabcolsep}{3.5pt}
\caption{LongMemEval-S methods with estimable API cost, sorted by accuracy (desc.). Token cost is USD per $10^{3}$ questions ($=\!1000\times$ per-question cost). SodaMem reports \emph{mean} cost to match baseline conventions; median is \$1.11/$10^{3}$Q ($\approx$14.6k tokens). ``Est.'' = priced from disclosed tokens; ``Meas.'' = author-reported / measured USD. SodaMem row highlighted.}
\label{tab:lme-cost}
\begin{tabular}{@{}>{\raggedright\arraybackslash}p{2.55cm}cl>{\raggedright\arraybackslash}p{2.35cm}rr@{}}
\toprule
\textbf{Method} & \textbf{Cite} & \textbf{Date} & \textbf{Model} & \textbf{Acc.} & \textbf{Cost/$10^{3}$Q} \\
\midrule
agentmemory V4 & \cite{agentmemory2026} & 2026-03 & Claude Opus 4.6 & 96.2\% & $\$60$ (est.) \\
Mem0 (2026 research) & \cite{mem02026research} & 2026-04 & Managed (GPT-4o est.) & 94.4\% & $\$22$ (est.) \\
\rowcolor{blue!12}
\leavevmode\bfseries SodaMem (ours)$^\dagger$ & --- & 2026-08 & \bfseries deepseek-v4-flash & \bfseries 92.8\% & \bfseries \$1.61 (meas.) \\
Cersei Full-context & \cite{cersei2026} & 2026-04 & Gemini 2.5 Flash & 87.6\% & $\$33$ (meas.) \\
Cersei Embed & \cite{cersei2026} & 2026-04 & Gemini 2.5 Flash & 86.6\% & $\$1.84$ (meas.) \\
Cersei Hybrid & \cite{cersei2026} & 2026-04 & Gemini 2.5 Flash & 86.3\% & $\$10$--$16$ (meas.) \\
AgentOS & \cite{agentos2026} & 2026-04 & GPT-4o & 85.6\% & $\$7.7$ (meas.) \\
LC GPT-5-mini & \cite{factmemcost2026} & 2026-03 & GPT-5-mini & 82.4\% & $\$29.3$ (meas.) \\
EmergenceMem Simple Fast & \cite{emergence2025,agentos2026} & 2025-06 & GPT-4o & 79.0\% & $\$46$ (meas.) \\
MemOS (eval set) & \cite{memoseval2025} & 2025-07 & GPT-4o-mini & 77.8\% & $\$0.33$ (est.) \\
TiMem & \cite{timem2026} & 2026-01 & GPT-4o-mini & 76.9\% & $\$0.31$ (est.) \\
Memobase & \cite{memoseval2025} & 2025-07 & GPT-4o-mini & 72.4\% & $\$0.35$ (est.) \\
MemOS (TiMem repro) & \cite{timem2026,memos2025} & 2026-01 & GPT-4o-mini & 68.7\% & $\$0.28$ (est.) \\
Mem0 (TiMem repro) & \cite{timem2026,mem02025} & 2026-01 & GPT-4o-mini & 65.0\% & $\$0.37$ (est.) \\
Zep (eval set) & \cite{memoseval2025,zep2025} & 2025-07 & GPT-4o-mini & 63.8\% & $\$0.36$ (est.) \\
Supermemory (eval set) & \cite{memoseval2025} & 2025-07 & GPT-4o-mini & 58.4\% & $\$0.18$ (est.) \\
MemoryOS & \cite{timem2026,memoryos2025} & 2026-01 & GPT-4o-mini & 58.0\% & $\$1.26$ (est.) \\
A-MEM & \cite{timem2026,amem2025} & 2026-01 & GPT-4o-mini & 55.4\% & $\$0.72$ (est.) \\
Fact-Mem0 (read) & \cite{factmemcost2026} & 2026-03 & GPT-5-mini & 49.0\% & $\$1.3$ (meas.) \\
MemU & \cite{memoseval2025} & 2025-07 & GPT-4o-mini & 38.4\% & $\$0.20$ (est.) \\
MemoryBank & \cite{timem2026,memorybank2023} & 2026-01 & GPT-4o-mini & 21.0\% & $\$2.21$ (est.) \\
\bottomrule
\end{tabular}

\vspace{2pt}
\begin{minipage}{\textwidth}
\footnotesize $^\dagger$Mean over 500 questions (planner+reader; excl.\ ingest/judge). \textbf{Median}: $\$1.11$/$10^{3}$Q $\approx$14.6k tokens/question---more representative of a typical query; the mean is pulled up by a long tail.
\end{minipage}
\end{table*}

\subsection{Setup and Cost Protocol}

\paragraph{SodaMem run.}
Entity-subject store-of-record (500 users, 235{,}840 facts). Planner, reader, and judge are \texttt{deepseek-v4-flash} with LongMemEval's official yes/no templates. Accuracy is $464/500$ ($92.8\%$; best of $N{=}3$; median $90.6\%$). End-to-end planner+reader \texttt{usage\_totals} (excluding ingest and judge), priced at Flash list rates ($\$0.14$ / \$0.0028 / \$0.28 per 1M for cache-miss / cache-hit input / output), yield a \textbf{mean} of $18{,}348$ tokens/question and \textbf{\$0.00161/question} (\$1.61 per $10^{3}$Q in Table~\ref{tab:lme-cost}). A long tail pulls the mean up: the \textbf{median} is $14{,}640$ tokens and \textbf{\$0.00111/question} (\$1.11 per $10^{3}$Q), $\approx$25\% lower, so a typical question is cheaper than the mean bill suggests. The same Flash model grades the run (self-grading); absolute accuracy may shift under an independent GPT-4o judge, but released hypotheses support re-evaluation and cost is judge-independent.

\paragraph{Baseline cost estimation.}
Author-reported USD (or \$/correct) is used when available (AgentOS~\cite{agentos2026}, Cersei~\cite{cersei2026}, Fact-Mem0 read~\cite{factmemcost2026}, EmergenceMem Simple Fast via AgentOS~\cite{emergence2025,agentos2026}). Otherwise we price disclosed tokens with the reported answering model: GPT-4o-mini $\$0.15/\$0.60$, GPT-4o $\$2.50/\$10$, Gemini 2.5 Flash $\$0.30/\$2.50$, Claude Opus 4.6 $\$5/\$25$, GPT-5-mini $\$0.25/\$2$ (per 1M tokens; $90\%/10\%$ in/out prior if undisclosed). TiMem Table~6~\cite{timem2026} and MemOS\_eval\_result~\cite{memoseval2025} report \emph{recalled context} length; we price that plus $\approx$200 output tokens as an answer-stage lower bound. Mem0's 2026 research mean tokens~\cite{mem02026research} are priced as GPT-4o under the same prior. Unless marked measured, costs are estimates---order-of-magnitude comparisons, not milli-dollar rankings. Frozen store fingerprints and usage totals accompany the $92.8\%$ artifact.

\subsection{Result Analysis}

\paragraph{Where SodaMem sits.}
At $92.8\%$ and mean \$0.00161/question ($\approx$18.3k tokens), SodaMem occupies a high-accuracy, mid-low-cost point (Figure~\ref{fig:cost-acc}); the median (\$0.00111; $\approx$14.6k) is more favorable for a typical query, so the plotted mean is a conservative reading of our own distribution. Two higher scores---agentmemory V4 at $96.2\%$~\cite{agentmemory2026} and Mem0 2026 at $94.4\%$~\cite{mem02026research}---sit roughly an order of magnitude to the right (\$0.06 and \$0.022 under our assumptions), reflecting Opus / GPT-4o-class generators rather than Flash. Unified GPT-4o-mini academic pipelines (TiMem, MemOS, Memobase, Zep~\cite{timem2026,memoseval2025}) are cheaper on the answer-stage lower bound but land at $\approx$58--78\%---well below our planner--reader loop.

\paragraph{Dominated quadrant and reader tier.}
The shaded region (cost $>$ mean \$0.00161 and accuracy $<$ $92.8\%$) contains Cersei Embed / Hybrid / Full-context~\cite{cersei2026}, AgentOS~\cite{agentos2026}, long-context GPT-5-mini~\cite{factmemcost2026}, EmergenceMem Simple Fast~\cite{emergence2025}, and MemoryBank under TiMem~\cite{timem2026,memorybank2023}---strictly worse (cost, accuracy) pairs even against our mean. Under the median (\$0.00111), MemoryOS and Fact-Mem0 read would enter as well. Public accuracy jumps often track reader upgrades (e.g., Mastra $84.23\%\rightarrow 94.87\%$ from GPT-4o to GPT-5-mini~\cite{mastraom2026}); SodaMem's claim is \textbf{near-frontier accuracy at Flash-tier spend}, undercutting Opus/GPT-4o high-score systems by $\approx$10--40$\times$ in estimated \$/question. Caveats: protocols and judges differ; recall-context pricing undercounts multi-call planners; ingest amortization varies. With those limits, Figure~\ref{fig:cost-acc} still shows a competitive accuracy band outside the high-cost frontier cluster, and strict dominance of several published points.

\paragraph{Limitations.}
This preprint reports a single store-of-record configuration under Flash self-grading; we do not claim a unified re-run of all baselines. Cost figures for many peers are reconstructed from disclosed tokens or author USD and should be read as order-of-magnitude. Ingest-time spend and timeline-resolution ablations are left for follow-up.

\section{Conclusion}

We presented \textbf{SodaMem}, an evidence-grounded temporal graph memory for LLM agents: typed FactEvents with provenance, temporal axes and supersession, hybrid retrieval, and a planner--reader answering loop. On LongMemEval-S, our store-of-record configuration reaches $92.8\%$ accuracy at mean \$0.00161 per question ($\approx$18.3k tokens; median \$0.00111 / $\approx$14.6k) with deepseek-v4-flash. Relative to public systems with estimable API cost, this point sits near the accuracy frontier while avoiding the Opus/GPT-4o high-spend cluster, and it strictly dominates several published (cost, accuracy) pairs. Remaining misses---especially temporal reasoning under self-grading---motivate session-anchored timeline resolution at ingest and independent re-judging of released answer hypotheses. We plan to release code, prompts, frozen store fingerprints, and the cost--accuracy compilation to support reproducible comparison.


\end{document}